\documentclass{ifacconf}

\usepackage{graphicx}      
\usepackage{natbib}        

\usepackage{amsmath}
\usepackage{amssymb}
\usepackage{commath}
\usepackage{mathtools}
\usepackage{graphicx}
\usepackage[width=.48\textwidth]{caption}
\usepackage{subcaption}

\usepackage[ruled,vlined,algo2e]{algorithm2e}
\usepackage{xcolor}
\usepackage{soul}
\begin{document}
\begin{frontmatter}

\title{Grounding-aware RRT* for Path Planning and Safe Navigation of Marine Crafts in Confined Waters\thanksref{footnoteinfo}} 

\thanks[footnoteinfo]{This research was sponsored by the Danish Innovation Fund, The Danish Maritime Fund, Orients Fund and the Lauritzen Foundation through the Autonomy part of the ShippingLab project, grant number 8090-00063B. The electronic navigational charts have been provided by the Danish Geodata Agency.\\
	\\
	\text~copyright~2021 the authors. This work has been accepted to IFAC for publication under a Creative Commons Licence CC-BY-NC-ND}

\author[First]{Thomas T. Enevoldsen} 
\author[First]{Roberto Galeazzi}

\address[First]{Automation and Control Group, Department of Electrical Engineering, Technical University of Denmark, DK-2800 Kgs. Lyngby, Denmark (e-mail: \{tthen,rg\}@elektro.dtu.dk).}

\begin{abstract}
The paper presents a path planning algorithm based on RRT* that addresses the risk of grounding during evasive manoeuvres to avoid collision. The planner achieves this objective by integrating a collective navigation experience with the systematic use of water depth information from the electronic navigational chart. Multivariate kernel density estimation is applied to historical AIS data to generate a probabilistic model describing seafarer's best practices while sailing in confined waters. This knowledge is then encoded into the RRT* cost function to penalize path deviations that would lead own ship to sail in shallow waters. Depth contours satisfying the own ship draught define the actual navigable area, and triangulation of this non-convex region is adopted to enable uniform sampling. This ensures the optimal path deviation. 
\end{abstract}


\end{frontmatter}

\section{Introduction}
According to a report by the European Maritime Safety Agency, incidents rooted within human error, such as collision, contact and grounding, account for $44\%$ of the total, with specifically grounding accounting for $13\%$ \citep{emsa2020}. \cite{uugurlu2015analysis} investigated grounding incidents caused by human error, and found that poor usage of the available equipment and poor assessment of the situation are key triggers. Decision support systems for collision and grounding avoidance can significantly improve the safety of navigation, by providing the human navigator with path alterations that are optimal with respect to the collision scenario and surrounding environment. 

Navigation in confined waters is challenging since collision avoidance must be weighed against the risk of grounding, possibly overestimated by navigators in search for more comfortable, deeper waters. Nonetheless, the collective navigation experience of seafarers across years of sailing may enable the creation of ``good seamanship'' models to inform the collision and grounding avoidance algorithms about best sailing practices. This paper researches a method to harvest such collective navigation experience and encode it into a probabilistic model describing seafarers' sailing behaviour in confined waters. Further, the paper specializes the RRT* path planning algorithm by integrating this probabilistic model with water depths information retrieved from the electronic navigational chart~(ENC) to compute optimal path deviations to deconflict collision scenarios.

\subsection{Related work}
Methods for quantifying and including relevant environmental information for safe navigation and collision avoidance within confined waters has been widely investigated. The review by \cite{huang2020ship} provides a broad insight into the typical components required for collision avoidance methods, with a distinction between motion prediction, conflict detection and resolution. 
\cite{Vagale2021,Vagale2021a} presented a detailed review of the path planning aspect of collision avoidance, discussing the potential advantages and challenges for autonomous surface vessels, as well as presenting various planning schemes.

\cite{chen2020global} used a binary occupancy grid along with the Fast Marching Method, and \cite{singh2018constrained} similarly with the A* algorithm.
\cite{bitar2020two} detailed a two-stage trajectory planner, where the initial step uses a discretized polygonal representation of the configuration space, such that a hybrid A* algorithm can compute an initial dynamically feasible trajectory.
\cite{xue2011automatic} generated potential functions from arbitrary polygons, as well as satellite images, for use with an Artificial Potential Fields algorithm.
Both \cite{candeloro2017voronoi} and \cite{niu2018energy} combine the use of Voronoi diagrams with real spatial data.
\cite{martinsen2020optimal} detailed the use of Constrained Delaunay Triangulation on land contours, to obtain an adjacency graph for trajectory refinement.
\cite{tsou2016multi} formed obstacles and Predicted Areas of Danger (PADs) by utilizing information from the Electronic Chart Display and Information System (ECDIS) and AIS, in order to plan using a genetic algorithm.
\cite{reed2016providing} and \cite{otterholm2019extracting} both detailed the use S-57 charts in the context of collision avoidance, and presented various approximations for the obstacles.\\

Sampling-based planning strategies are prime-candidates for computing both feasible and optimal paths within the space such as those defined by the non-convex depth contours. Since its inception, the Rapidly-exploring Random Trees (RRT) algorithm \citep{lavalle1998rapidly} has played an important role in sampling-based strategies, with \cite{karaman2011sampling} introducing the asymptotically optimal RRT*. 
Within sampling-based collision avoidance schemes for marine crafts, \cite{Chiang2018} detailed the use of a non-holonomic RRT, modelling the land masses as arbitrary polygons. \cite{zaccone2021} presented a COLREGs compliant RRT*, which minimized path length, curvature and repulsion from objects. \cite{enevoldsen2021colregs} presented the COLREGs-Informed RRT*, which directly samples COLREGs compliant trajectories for single vessel encounters in open-waters.

\subsection{Novelty and contribution} 
This paper proposes a novel marine-oriented RRT* path planning algorithm for collision and grounding avoidance of vessels sailing in confined waters. Current sampling-based algorithms adopted within the marine domain, such as RRT*, consider simplified or artificial representations of the environment. In contrast this contribution operates directly on environment information (e.g. depth contours) extracted from the real electronic navigational chart. We introduce the concept of \emph{collective navigation experience} to describe the best practices of seafarers while sailing in a given restricted sea region. We propose to use multivariate kernel density estimation on historical AIS data to generate probabilistic models that describe the seafarers' behaviour in steering vessels of different draught. We define the planning region by extracting from the ENC depth contours satisfying the vessel's draught requirements, and we propose triangulation of such non-convex polygons as a mean to achieve uniform sampling of the space. This ensures optimality of the sought path deviation. The probabilistic model is exploited in the RRT* cost function to sway the path planner towards deconflicting path alterations that favor comfortable navigation, as seen from the eyes of human navigator, while still exploring the whole navigable area as described by the ENC.

\section{Preliminaries}
\subsection{Vessel assumptions}
\subsubsection{Own ship}This study concerns path planning for large vessels with maneuvering restrictions due to their draught and minimum required turning radius, such as ro-ro vessels, container feeders, bulk carriers, etc.
It is assumed that a track control system is responsible to steer the vessel based on a sequence of waypoints $W_i = (N_i, E_i , R_i)$, where $N_i$ is the North position, $E_i$ is the East position, and $R_i$ is the radius of acceptance. The manoeuvring capabilities (e.g. minimal turning radius) of own ship are accounted for when generating a feasible waypoint sequence. 

\subsubsection{Target vessels} Target vessels are characterized by means of a comfort zone described by an ellipse whose dimensions are related to the ship length $L$, with $8L$ and $3.2L$ defining major and minor axis, respectively \citep{hansen2013empirical}.

\subsection{COLREGs description and compliance}
\begin{figure}[tb]
\centering
\includegraphics[width=0.45\textwidth]{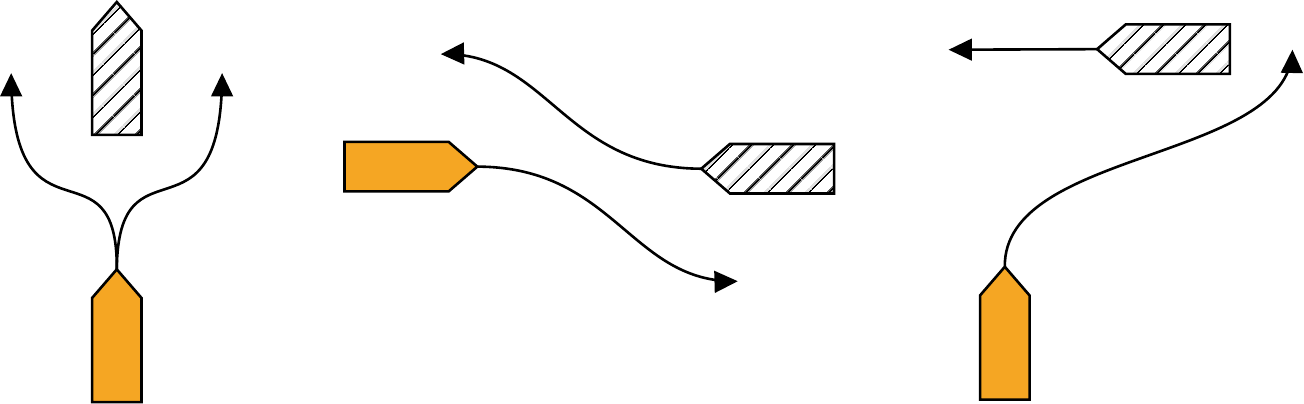}
\caption{Encounters described by COLREG rules 13-15, i.e. overtaking, head-on and crossing scenarios. The solid coloured vessel is in a give-way situation.}\label{R13_15}
\end{figure}

Any collision avoidance system must adhere to the ``rules-of-the-road'', namely the COLREGs. The most frequently considered rules are Rules 13-15, which deal with overtaking, head-on and crossing scenarios, respectively. The rules are defined for single-vessel encounters, assuming both vessels are power-driven. Figure~\ref{R13_15} visualizes the before-mentioned rules, where the vessel in give-away situation must yield for the stand-on vessel. For Rules 14 and 15, the give-way vessel must pass on the starboard side of the stand-on vessel, whereas both a port and starboard passing for Rule 13 is valid. 
The relative bearing between own ship and target vessel determines the collision scenario at hand. 
For further details regarding the relevant COLREGs, interpretation of the relative bearing, and assessment of the collision scenario see~\citep{enevoldsen2021colregs}.

\subsection{Electronic Navigational Charts}
\begin{figure*}[htb]
     \centering
     \begin{subfigure}[b]{0.52\columnwidth}
         \centering
         \includegraphics[width=\textwidth]{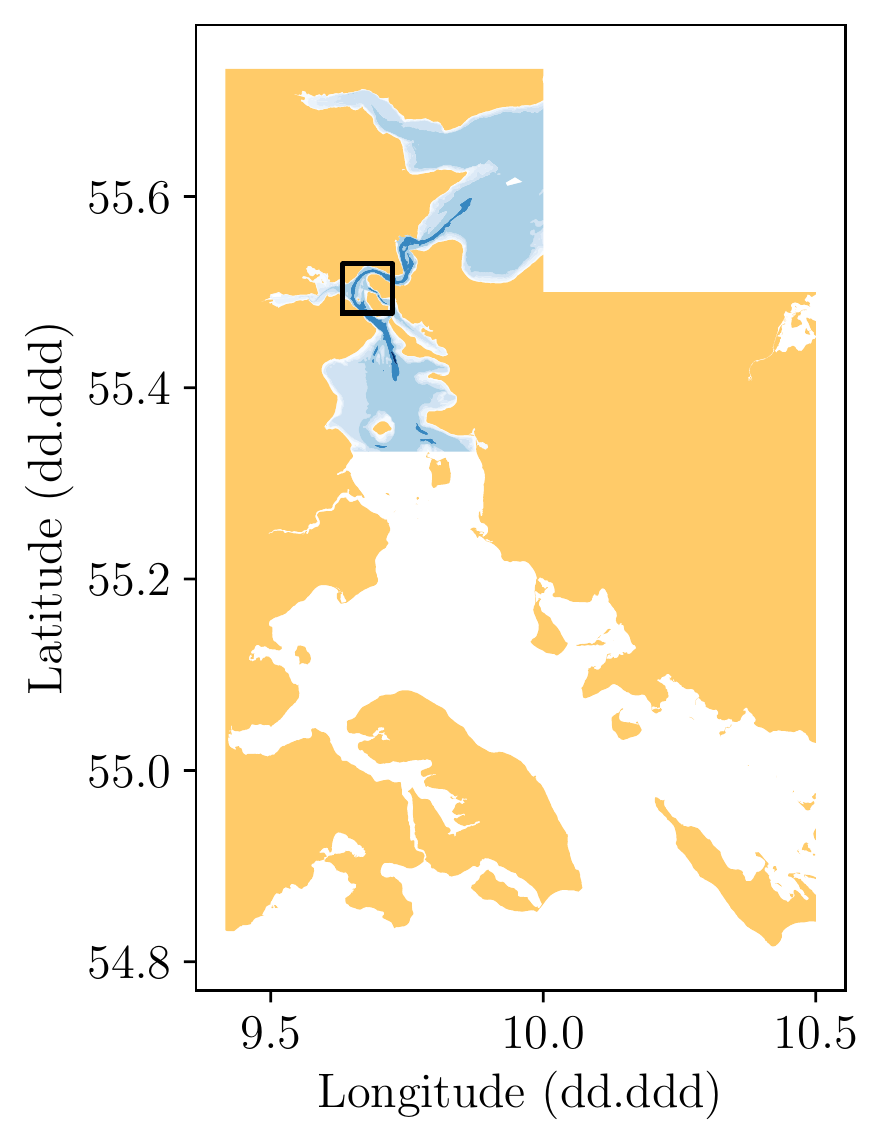}
         \caption{Geofiltered (black)}
         \label{fig:uncropped_smallbelt}
     \end{subfigure}
     \hfill
     \begin{subfigure}[b]{0.72\columnwidth}
         \centering
         \includegraphics[width=\textwidth]{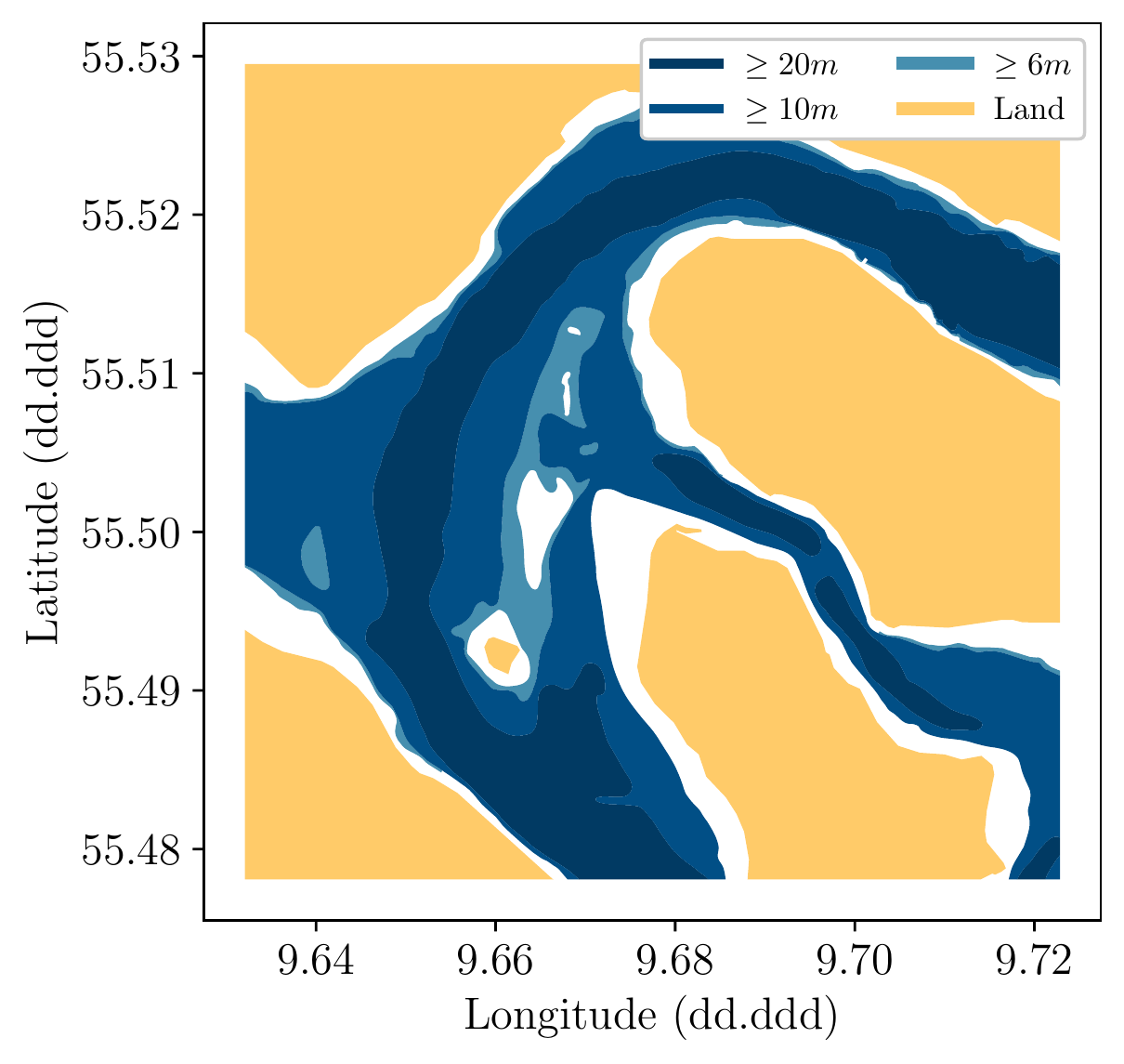}
         \caption{Cropped contours}
         \label{fig:cropped_smallbelt}
     \end{subfigure}
     \hfill
    \begin{subfigure}[b]{0.72\columnwidth}
         \centering
         \includegraphics[width=\textwidth]{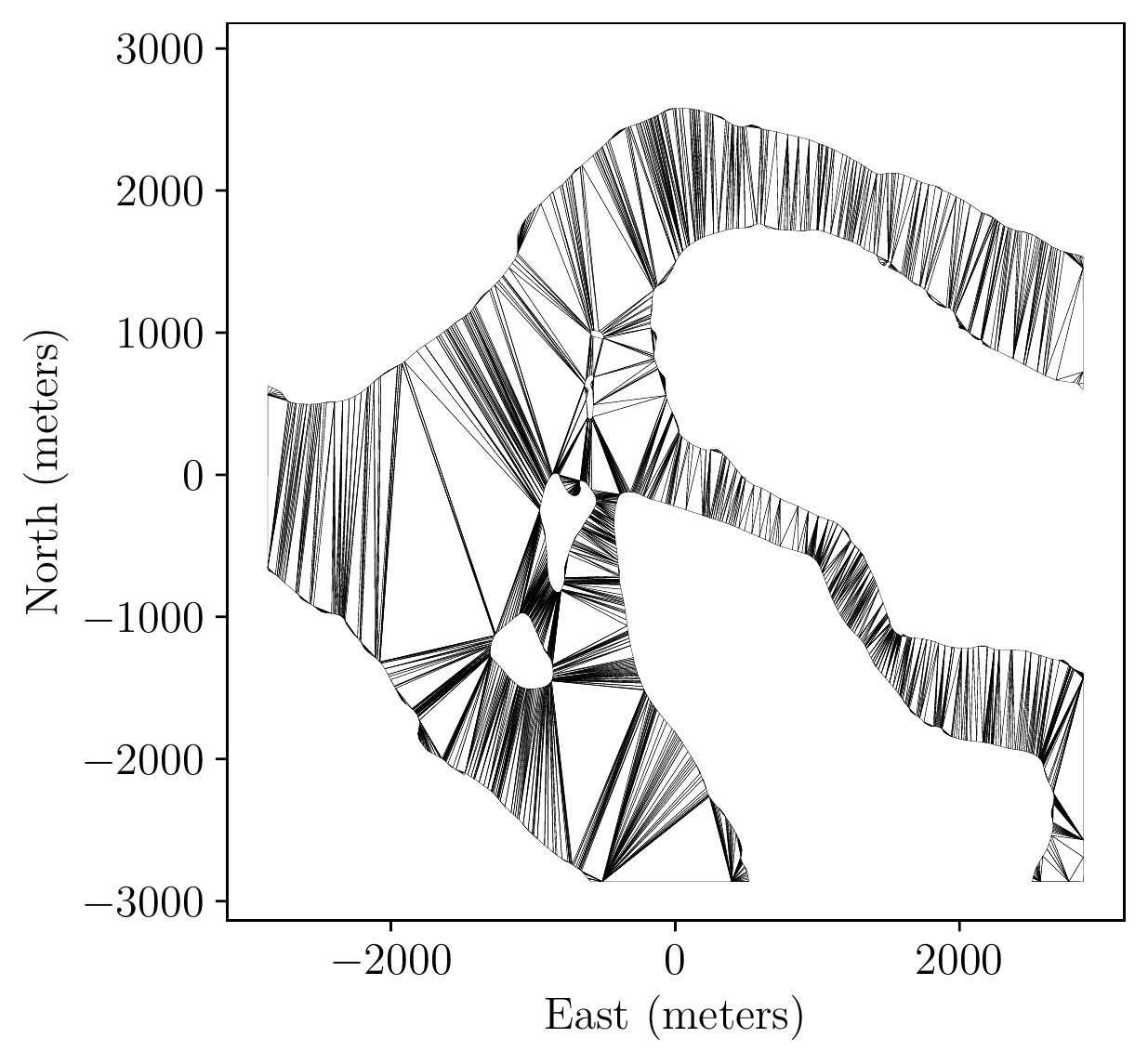}
         \caption{Triangulated feasible depths}
         \label{fig:triangulated_smallbelt}
     \end{subfigure}
        \caption{Extracting data from the S-57 ENCs using a pre-defined geofilter (black square in Fig.~\ref{fig:uncropped_smallbelt}), and subsequently manipulating the polygons in order to extract contours with a specified minimum water depth (Fig.~\ref{fig:cropped_smallbelt}).}
        \label{fig:enc_data_comparison_smallbelt}
\end{figure*}

The electronic chart display and information system is a compliant, digital alternative to paper nautical charts. The ECDIS serves the navigator with important information, typically contained within the conventional paper charts, as well as information from RADAR and AIS. The Electronic Navigational Charts (ENCs) must conform to the S-57 standard in order to be used with the ECDIS. Within the S-57 charts various information layers exist, containing data about land and depth contours, buoys identifiers and placement, recommended navigational tracks, dredges, restricted areas, etc. 
The information contained within the ENC is fundamental for navigating in confined waters, since it directly encodes geographical areas which are feasible with respect to the draught of a given vessel. Therefore, the information layer of the ENC utilized for this particular study is that one containing the depth contours, namely \texttt{DEPARE}. For visualization purposes the land masses contained within \texttt{LNDARE} are also extracted. 
Through software, all depth contours intersecting with a geofilter are selected and subsequently cropped (Fig.~\ref{fig:uncropped_smallbelt}), resulting in a collection of polygons described by their geodetic coordinates, minimum and maximum water depths. Figure~\ref{fig:cropped_smallbelt} visualizes the cropped depth contours with minimum water depth greater than 6m (white areas are waters shallower than 6m).

\section{Collective Navigation Experience}

\subsection{Multivariate Kernel Density Estimation}
Kernel density estimation belongs to the family of non-parametric methods for estimating an unknown probability density function based on a finite data sample.

Let $\mathbf{X}_i = (X_{i1}, X_{i2}, \ldots, X_{ip})^\top$, $i= 1,2, \ldots, n$, be a sequence of independent and identically distributed $p$-variate random variables drawn from an unknown density $f$. Then, the general form of the $p$-dimensional multivariate kernel density estimator is~\citep{gramacki2018nonparametric}
\begingroup\belowdisplayskip=\belowdisplayshortskip
\begin{equation}
    \hat{f}(\mathbf{x},\mathbf{H}) = \frac{1}{n}\sum_{i = 1}^n|\mathbf{H}|^{-1/2}K(\mathbf{H}^{-1/2}(\mathbf{x} - \mathbf{X}_i))
\end{equation}\endgroup
where $\mathbf{H} = \mathbf{H}^\top > 0$ is the non-random $p\times p$ bandwidth matrix, and $K(\cdot)$ is the unscaled kernel. 

The bandwidth matrix and the kernel determines the degree of smoothness of the estimated density, which in fact inherits all the smoothness properties of the underlying kernel. Probability density functions are generally adopted as kernel functions. This implies that the area under the kernel must be equal to one, and that the kernel function is always positive. A commonly adopted kernel is the multivariate normal density
\begin{equation}
    K(\mathbf{z}) = \frac{1}{\sqrt{(2\pi)^p}}\exp{\left(-\frac{1}{2}\mathbf{z}^\top\mathbf{z}\right)}
\end{equation}
The bandwidth matrix $\mathbf{H}$ determines both the amount and orientation of the smoothing. In the most general case of an unconstrained bandwidth, there are $n(n+1)/2$ parameters to be determined in order to achieve the desired level of smoothing. Due to the complexity of solving the unconstrained case, it is common to constrain the bandwidth matrix such that $\mathbf{H} = h^2\mathbf{I}_p$ with $h>0$ or $\mathbf{H} = \mathrm{diag}\{h_1^2,h_2^2 \ldots, h_p^2\}$ with $h_i >0$. In these cases there is a wide array of methods that can be applied to estimate the values of the parameters $h_i$ from the available data \citep[Chapter 4]{gramacki2018nonparametric}.
Estimation problems addressing large data sets with high dimensionality show a computational complexity as high as $O(n^2)$, if a naive evaluation of the kernel function $K(\cdot)$ is performed. Several methods have been developed to accelerate the computation of the kernel functions \citep{raykar2010fast}, and one of the most effective applies the Fast Fourier Transform (FFT) \citep{silverman1982algorithm,gramacki2017fft}.

\subsection{Automatic Identification System (AIS)}

The overall collection of historical AIS data is represented as the set $\mathcal{A}$, whose elements are messages $\boldsymbol{m}_i$ 
\begin{equation}
    \boldsymbol{m}_i = \begin{bmatrix}
        \text{MMSI}_i & t_i & \text{SOG}_i & D_i &\lambda_i & \phi_i
    \end{bmatrix}
\end{equation}
containing the following data entries: the Maritime Mobile Service Identity $\text{MMSI}_i$, the timestamp $t_i$, the speed-over-ground $\text{SOG}_i$, the vessel draught $D_i$, the vessel position in geodetic coordinates (latitude $\lambda_i$ and longitude $\phi_i$).
The generation of the collective navigation experience for a specific geographical region is based on a subset of $\mathcal{A}$, which contains data pertaining to vessels sailing in such region. Let introduce the geodetic region of interest $\mathcal{R} = \{ (\lambda_i, \phi_i) \mid \lambda_{min} \leq \lambda_i \leq \lambda_{max} \land \phi_{min} \leq \phi_i \leq \phi_{max}\}$, and the speed range of interest $\mathcal{V} = \{\mathrm{SOG}_i \mid \mathrm{SOG}_{min} < \mathrm{SOG}_i < \mathrm{SOG}_{max}\}$. Then the set containing the data relevant to the estimation of the collective navigation experience for vessels with draught larger than or equal to a given draught $\bar{D}$ is defined as
\begin{equation}
    \mathcal{N}_{\!\bar{D}} \triangleq \left\{\boldsymbol{m}_i \in \mathcal{A} \mid (\lambda_i, \phi_i) \in \mathcal{R} \land \mathrm{SOG}_i \in \mathcal{V} \land D_i \geq \bar{D} \right\}
\end{equation}
The set $\mathcal{V}$ guarantees that only AIS messages from marine vessels are considered, leaving out from the set $\mathcal{N}_{\!\bar{D}}$ data referring to drifting buoys and fast moving rescue vehicles.

\subsection{Learning navigators' best practices in confined waters}
Navigating within confined waters presents a plethora of new variables for navigators to consider, including the trade-off between voyage duration and safe passage. Safe navigation in complex waters is heavily dependant on the traffic within said waters, and the constraints presented by shallow waters. These trade-offs occurring within Danish waters have been documented through a national database of AIS messages. By computing multivariate KDEs using the past positions of vessels within the region of interest, the collective navigation experience can be quantified in a probabilistic sense.

\begin{figure}[tb]
\centering
\includegraphics[width=0.48\textwidth]{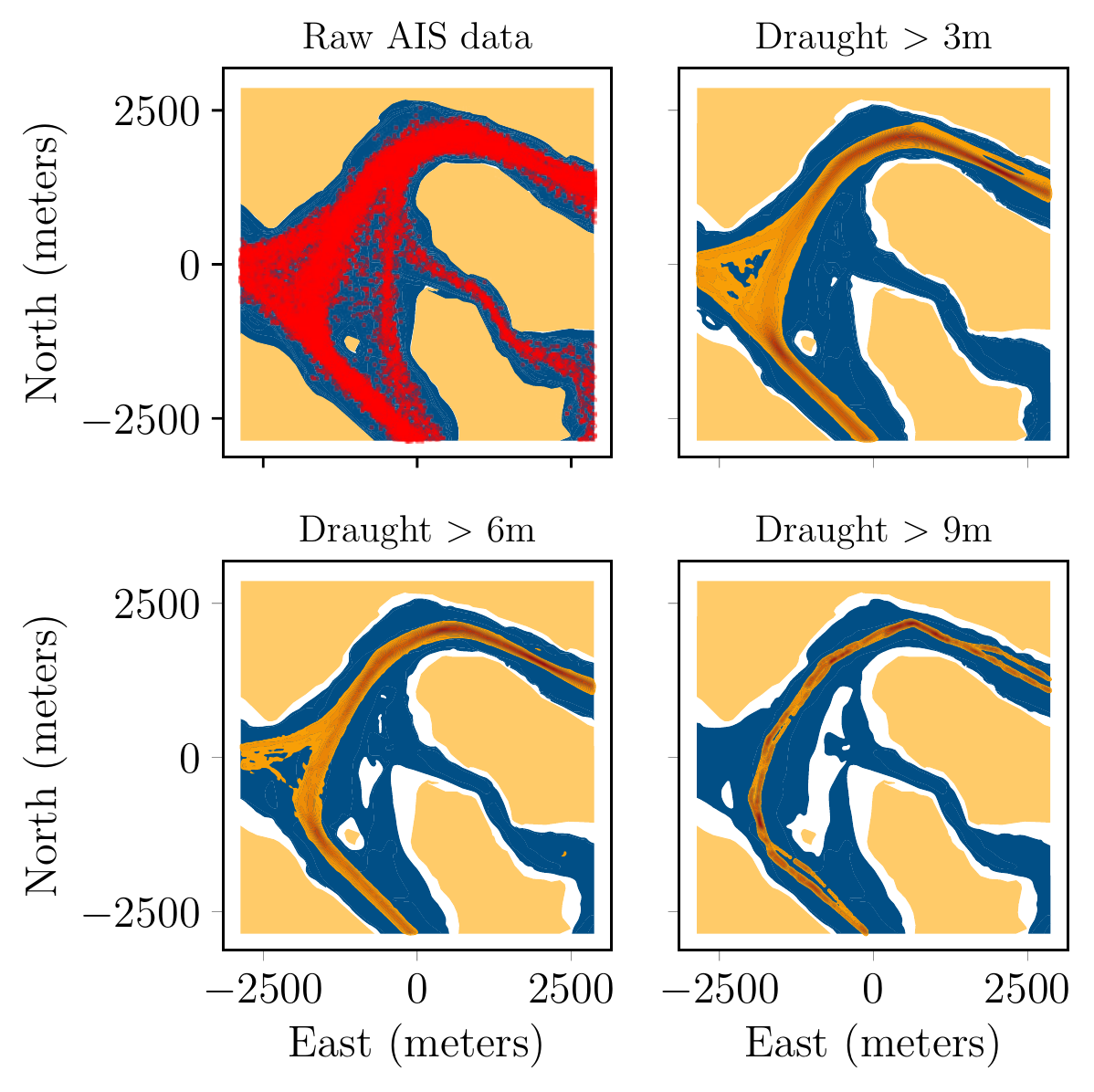}
\caption{Various KDEs within the Little Belt area of Denmark, for different draughts and depths.}\label{fig:KDE_draught_comp}
\end{figure}
The top left corner of Fig.~\ref{fig:KDE_draught_comp} shows an example of AIS data for the region of interest $\mathcal{R} = \{ (\lambda_i, \phi_i) \mid 55.48^\circ \leq \lambda_i \leq 55.53^\circ \land 9.64^\circ \leq \phi_i \leq 9.72^\circ\}$, corresponding to the Little Belt area in Denmark. The data refers to vessels navigating in the speed range of interest $\mathcal{V} = \{\mathrm{SOG}_i \mid 0.5\text{kn} < \mathrm{SOG}_i < 50\text{kn}\}$. Different KDEs can then be estimated by creating the set $\mathcal{N}_{\!\bar{D}}$ for different values of $\bar{D}$. For the estimation process we utilized the multivariate normal density as kernel function, and the constrained bandwidth model $\mathbf{H} = h^2\mathbf{I}_p$ where $h$ was hand-tuned. The actual estimation of the kernel density was performed leveraging the Python toolbox \texttt{KDEpy} \citep{odland2018kdepy}; specifically the FFT-based KDE scheme was utilized for its superior computational speed on the large AIS data set. Figure~\ref{fig:KDE_draught_comp} details the resulting KDEs estimated on three sets $\mathcal{N}_{\!\bar{D}}$ for increasing value of $\bar{D}$, and the underlying ENC data corresponds to waters whose minimal depth is greater than the specified draught. By excluding parts of the data set based on draught restrictions, the overall encoded behaviour from previous navigators changes. An important thing to note in this particular study-case is that, despite there is a direct northbound path, the majority of density moves North-West, before continuing North-East. This indicates that through past experiences, the navigators deem the passage directed North less safe, due to the heavily restricted waters, compared to prolonging the journey and increasing the overall safety margin. This shows that the probabilistic representation contains valuable information regarding the safe navigation of confined water areas. As observed, past navigators consistently practice navigational safety, which is evident within the historical data, where the majority of the density maintains a safe distance from shallow waters and other waters which are associated with higher navigational risks.

\section{Grounding-aware RRT*}
\subsection{Problem definition}
The generalized optimal planning problem considered in this paper is defined in a similar fashion to \citep{Gammell2014}. Let $\mathcal{X} \subseteq \mathbb{R}^{n}$ be the state space, which is divided into two subsets, $X_{\text{free}}$ and $X_{\text {obs}}$, with $X_{\text{free}}=\mathcal{X} \backslash X_{\text{obs}}$. 
The states within $X_{\text{free}}$ contain all states that are feasible with respect to maneuvering constraints, COLREGs compliance, collision and grounding. 
Let $\mathbf{x}_{\text{start}} \in X_{\text{free}}$ be the initial state at $t=0$ and $\mathbf{x}_{\text{goal}} \in X_{\text{free}}$ be the desired final state. 
Let $\sigma:[0,1] \mapsto \mathcal{X}$ be a sequence of states constituting a found path and $\Sigma$ be the set of all nontrivial and feasible paths. 
The objective is then to find the optimal path $\sigma^{*}$, which minimizes a cost function while connecting $\mathbf{x}_{\text{start}}$ to $\mathbf{x}_{\text{goal}}$ through $X_{\text{free}}$,
\begin{equation}
\begin{split}
\sigma^{*}=\underset{\sigma \in \Sigma}{\arg \min }\left\{c(\sigma) \mid \right. \sigma(0)=\mathbf{x}_{\text {start }},\, &\sigma(1)=\mathbf{x}_{\text {goal }}, \\
\forall s \in[0,1],\, &\sigma(s) \in \left. X_{\text {free }}\right\}.
\end{split}
\end{equation}
We propose a cost function that trades off between two performance indexes: the path length and the grounding risk. The latter is defined as the complementary of the normalized kernel density estimate, and represents the discomfort of fellow navigators when manoeuvring vessels too close to shallow waters. The cost function reads as
\begin{equation}
c(\sigma) = w_1(1 - \bar{F}(\sigma)) + w_2l(\sigma)\label{eq:costfunc}
\end{equation}
with scalar weights $w_{i}$, the over all path length $l(\sigma)$, and
\begin{equation}
\bar{F}(\sigma) = \hat{f}(\sigma, \mathbf{H}) / \max \left(\hat{f}(\mathbf{X}, \mathbf{H})\right)
\end{equation}
is the normalized kernel density along the path $\sigma$.
\begin{rem}
The simple structure of the cost function \eqref{eq:costfunc} aims at emphasizing the role of the collective navigation experience encoded in the estimated kernel density towards the generation of the path deviation during a potential collision scenario. However, the RRT* offers a general framework where the cost function can include an arbitrary number of performance indexes focusing on e.g., speed loss, energy consumption, fuel consumption, etc.
\end{rem}

\subsection{RRT*}
\IncMargin{1em}
\begin{algorithm2e}[tb]
\small
    \LinesNumbered
	\SetKwInOut{Output}{Given}
	\Output{$x_{start}$, $x_{goal}$}
	\SetKw{Given}{$x_{\text{start}}$, $x_{\text{goal}}$}
	$V \leftarrow \{x_{start}\}$,$\quad E \leftarrow \emptyset$,$\quad \mathcal{T} = (V,E)$\;
	\For{\textup{i = $1\dots N$}}{
		$x_{rand} \leftarrow$\texttt{Sample($x_{goal}$)}\;
		
		$x_{nearest} \leftarrow$ \texttt{NearestNode($\mathcal{T}$,$x_{rand}$)}\;
		$x_{new} \leftarrow$ \texttt{ExtendTowards($x_{nearest}$,$x_{rand}$)}\;
		\If{\texttt{\upshape Feasible($x_{nearest},x_{new}$)}}{
		    $X_{near} \leftarrow$ \texttt{Near($\mathcal{T}$,$x_{new}$,$r$)}\;
			$x_{min} \leftarrow$ \texttt{BestParent}($X_{near}, x_{nearest}, x_{new}$)\;
			$\mathcal{T} \leftarrow$\texttt{InsertNode}($\mathcal{T}, x_{new}$)\;
			$\mathcal{T} \leftarrow$\texttt{ReWire}($\mathcal{T}, X_{near}, x_{min}, x_{new}$)\;
			
		}
		
		}
		\Return $\mathcal{T}$\;
	\caption{RRT*}\label{mainRRT}
\end{algorithm2e}
\DecMargin{1em}
The underlying RRT* algorithm is a marine-oriented variant of the general algorithm presented by \cite{karaman2011sampling}. Algorithm~\ref{mainRRT} describes the main structure of the general RRT* planner.
The routine consists of growing a tree $\mathcal{T} = (V,E)$, where $V$ and $E$ represent the sets of nodes and edges, respectively. A given node $x$ represents a particular state at time $t$, containing the North-East position of the vessel, with the connecting edges containing cost and constraint information.
The algorithm is modified to incorporate the cost function \eqref{eq:costfunc}. \texttt{Feasible($x_{nearest},x_{new}$)} ensures that only feasible nodes are added to the tree, and therefore rejects node sequences that are either non-COLREGs compliant, outside the bounds of $X_{\text{free}}$ or violates the remaining constraints (such as the minimum turning radius). 
Due to availability of depth contours within the ENC, which directly encode a draught dependant representation of $X_{\text{free}}$, the proposed sampling strategy is tailored in order to leverage the non-convex polygons that describe the navigable regions. The following section, details the proposed uniform sampling scheme of general non-convex depth contours.

\subsection{Sampling non-convex depth contours}
The depth contours extracted from the ENC are represented by arbitrary non-convex polygons. To achieve the uniform sampling of this space, thereby ensuring optimality of the sought path, the Constrained Delaunay Triangulation (CDT) \citep{chew1989constrained} is applied to decompose the non-convex polygon into a finite set of triangles. CDT guarantees that all the obtained triangles are contained within the perimeter of the polygon. Figure~\ref{fig:triangulated_smallbelt} shows CDT applied to the polygon data obtained from the depth contours in Fig.~\ref{fig:cropped_smallbelt}. The uniform sampling of the obtained triangles is a two-step procedure: first there is the uniform selection of a triangle, weighted by area; then the uniform sampling of the selected triangle takes place. Given any triangle composed by vertices $(A,B,C)$, a uniformly sampled point $P$ within the triangle is then described by
\begin{equation}
P=\left(1-\sqrt{r_{1}}\right) A+\sqrt{r_{1}}\left(1-r_{2}\right) B+\sqrt{r_{1}} r_{2} C
\end{equation}
where $r_1 \sim \mathcal{U}(0,1)$ and $r_2 \sim \mathcal{U}(0,1)$ \citep{osada2002shape}.The polygon geometries contained within the ENC are represented at a very a high resolution. Hence it is recommended to simplify the polygons using a line simplification method,  such as the Douglas-Peucker algorithm \citep{douglas1973algorithms}.

\begin{rem}
The standard RRT* algorithm performs sampling in a rectangular region surrounding the space where the path needs to be planned. For navigation in confined waters, the sampling of such space is highly inefficient with respect to finding the optimal path. This is because many samples are likely to fall in correspondence of land masses and shallow waters. Rejection sampling is therefore applied to retain only the valuable samples. On the contrary, sampling the triangular regions within the perimeter of the depth contours is highly efficient since all the samples in such region contributes to improve the found solution.
\end{rem}

\section{Results}
\begin{figure}[tb]
     \centering
    \begin{subfigure}[b]{1\columnwidth}
         \centering
         \includegraphics[width=0.99\textwidth]{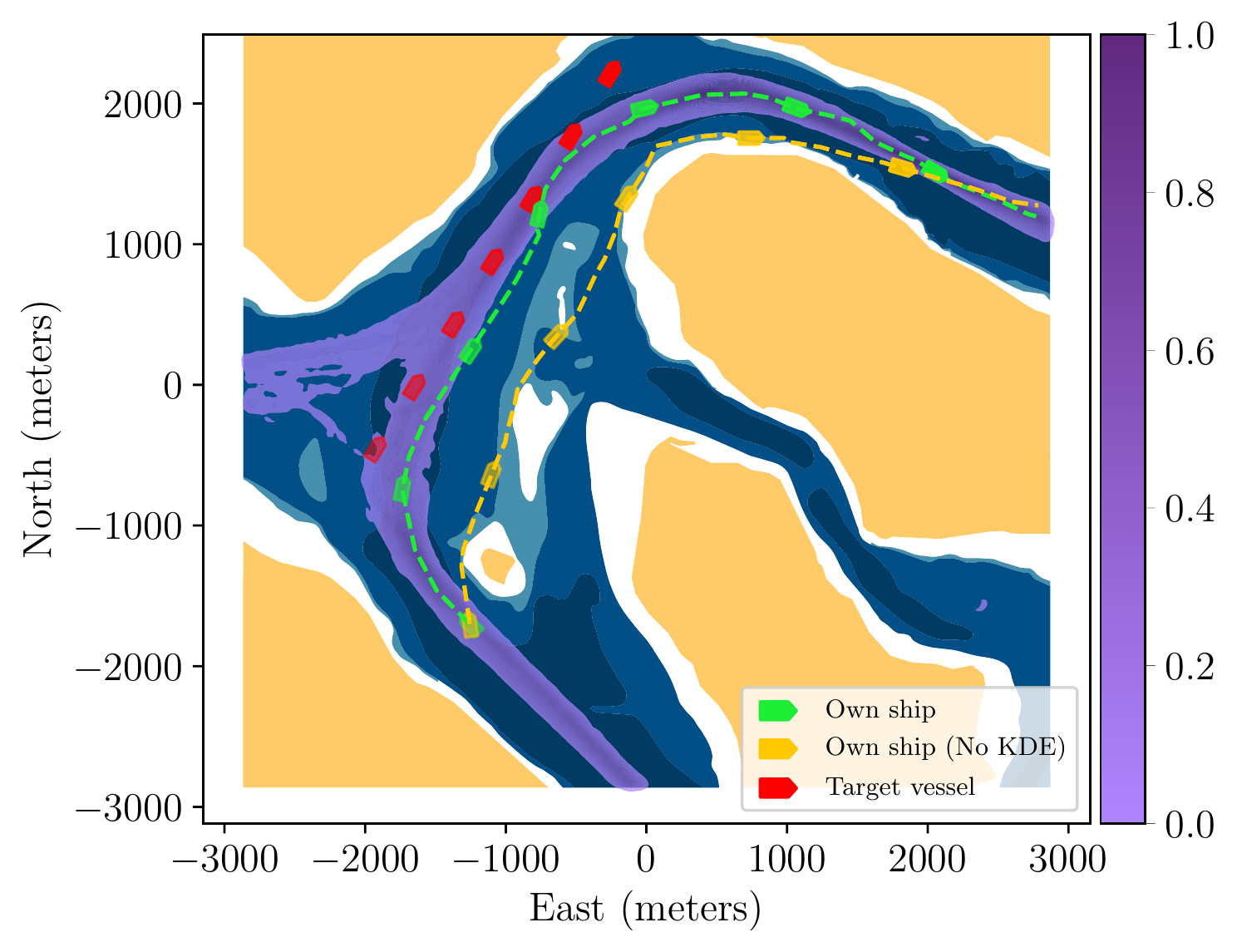}
         \caption{Overtaking scenario in the waters around Skærbæk, Fænø and Middelfart, at the Little Belt area in Denmark.}
         \label{fig:results_smallbelt}
     \end{subfigure}
     \hfill
    \begin{subfigure}[b]{1\columnwidth}
         \centering
         \includegraphics[width=0.99\textwidth]{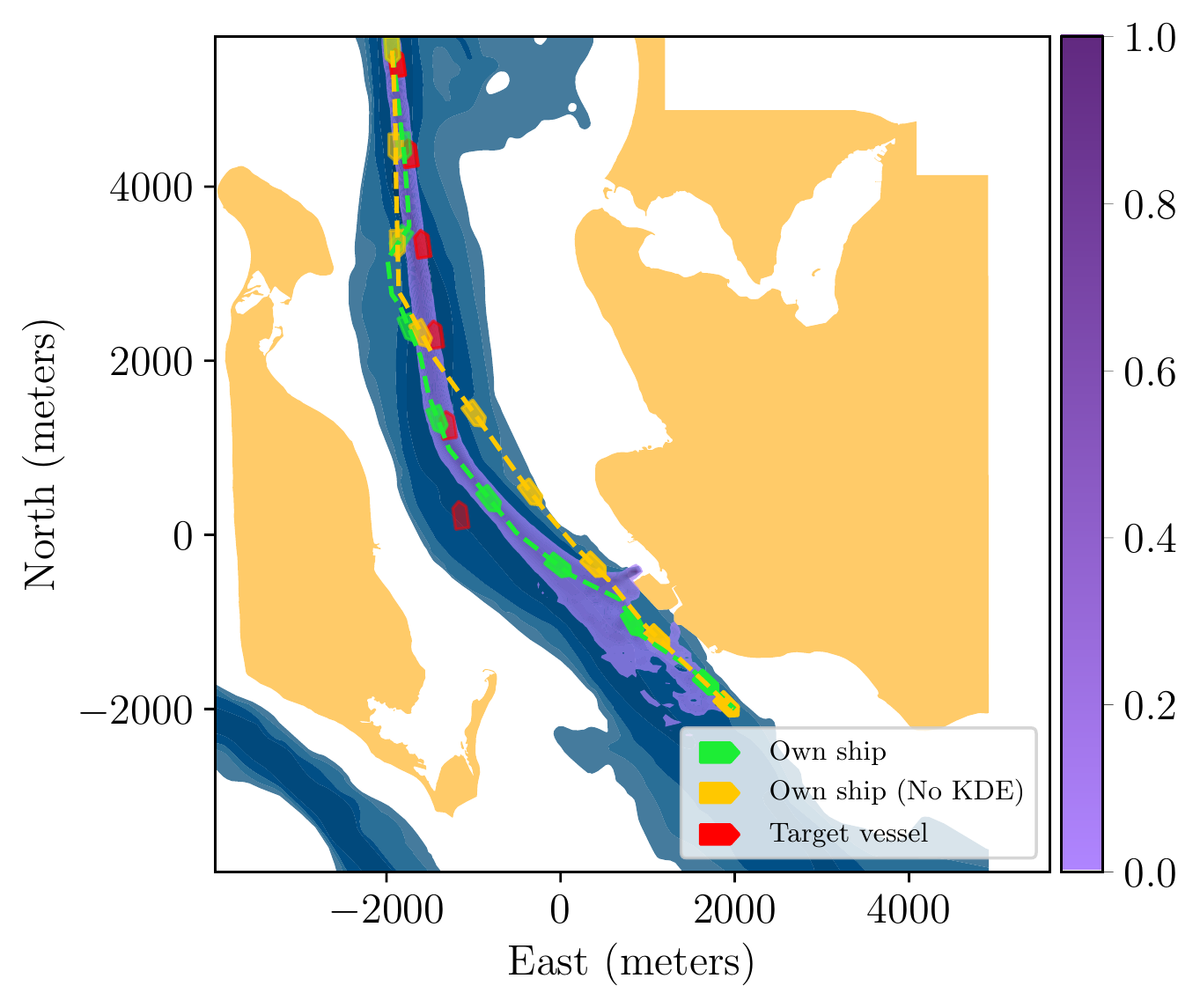}
         \caption{Head-on scenario in the waters between Agersø and Stignæs, at the Great Belt area in Denmark.}
         \label{fig:results_agersoe}
     \end{subfigure}
        \caption{Two scenarios for single-vessel encounters in confined water. The visualized depth contours are waters deeper than 6m and indicate the feasible area for own ship to traverse; white areas indicate shallow waters.}
        \label{fig:result_plots}
\end{figure}
The grounding-aware RRT* planner is evaluated on two confined water scenarios. Each scenario includes a single-vessel encounter, where own ship must perform an evasive manoeuvre to avoid collision. The planning scheme receives the position of own ship as the starting configuration, and selects a point along the nominal trajectory as the goal configuration. The resulting trajectories for both scenarios are visualized in Fig.~\ref{fig:result_plots}. Two different deviations, with differing objectives, were computed for each scenario. For the first objective, the weights of the cost function \eqref{eq:costfunc} were selected to emphasise the value provided by the learned behaviour from the AIS data, and for the second the weights were selected to only consider the minimization of the path elongation (i.e. $w_1 = 0$).

In the first scenario, Fig.~\ref{fig:results_smallbelt}, own ship is travelling northbound through the Little Belt area when it encounters a target vessel also northbound, but travelling at a lower speed. The encounter with the target vessel is identified as an overtaking scenario, using the relative bearing. The second scenario, Fig.~\ref{fig:results_agersoe}, features own ship heading South down a narrow channel at the Great Belt area. The grounding-aware RRT* planner is initiated due to an incoming target vessel that, based on the relative bearings, is identified as a head-on encounter. This implies that own ship must yield. 

In both study-cases, the grounding-aware RRT* path planner computes a COLREGs compliant path alteration, which ensures both the collision avoidance and the safe navigation across confined waters. The computed path alterations clearly show how the planner negotiates between the collective navigation experience embodied by the KDE and the water depth information provided by the ENC. The former attracts the path of own ship to traverse waters that have been heavily navigated in the past by other vessels of equal draught; the latter pushes the path to cross deep enough waters closer to shore to ensure that the comfort zone of the target vessel is never violated during the evasive manoeuvre. The combined effect is a path deviation that concomitantly achieves collision and grounding avoidance while safeguarding the navigation comfort of both own ship and target vessel. The two additionally computed deviations which seek to minimize the path length, also provide COLREGs complaint path alterations, however both disregard the risks associated with traversing possibly too shallow waters. This results in a more risky passage due to both potential grounding risks and the limited space for maneuvers given additional incoming vessels.

\begin{figure}[tb]
\centering
\includegraphics[width=0.48\textwidth]{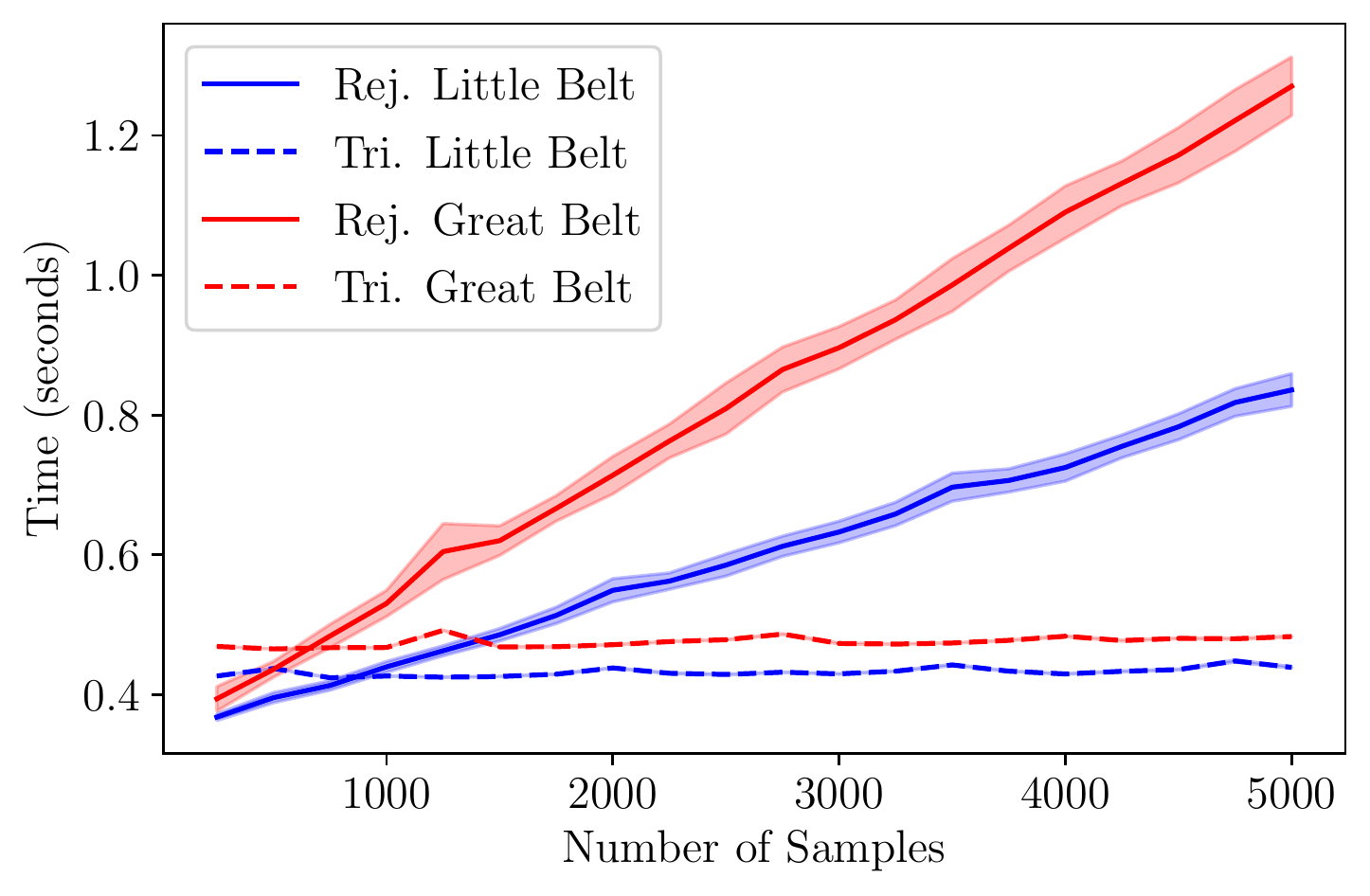}
\caption{Performance comparison between the rejection sampling scheme (Rej.) and the proposed scheme which samples the triangulated depth contours (Tri.). The area ratio between the rectangular region and the depth contours for the rejection sampling method is approximately $0.2677$ and $0.4065$ for the Great Belt and Little Belt, respectively. The transparent bands indicate the $3\sigma$ confidence interval.
}\label{fig:sampling_performance}
\end{figure}
\begin{table}[hb]
\begin{center}
\caption{Statistical performances in seconds related to the configuration of sampling spaces for the Great Belt (GB) and Little Belt (LB) scenarios.}\label{tb:config_ss}
\begin{tabular}{c|cccc}
 Scheme & GB($\mu$) & GB ($\sigma$) & LB ($\mu$) & LB ($\sigma$) \\\hline
Rejection & 0.3501 & 0.0087 &0.3469&0.0063 \\
Triangulation & 0.4730 & 0.0115 &0.4299&0.0108
\end{tabular}
\end{center}
\end{table}
The paper proposed using triangulation of the depth contours for computing valid samples in scenarios with confined waters. Its performance is compared against a standard sampling approach based on a rectangular region and rejection sampling. The comparison is performed using Monte Carlo simulations. Figure~\ref{fig:sampling_performance} illustrates the computational time in seconds as the number of valid samples increases: it is evident that the proposed triangulation scheme outperforms the baseline rejection sampler. Additionally, the performance related to configuring the sampling spaces is provided in Table~\ref{tb:config_ss}.

For both scenarios, the triangulation scheme is initially at a disadvantage, since on average it takes longer to extract the contours and compute the triangles, compared to simply extracting the depth contours as required by the rejection sampler. It should be noted, that due to variations in the polygonal data between the two scenarios, the preprocessing duration differs between the two presented scenarios. 
However, once the configuration of the sampling spaces is finalized, the proposed triangulation based sampling scheme significantly outperforms the rejection sampling scheme at higher sample quantities. The ability to rapidly compute valid samples is key for optimal sampling-based algorithms and the speed in which they converge to the respective optimal solution~\citep{Gammell2014,enevoldsen2021colregs}.
Noteworthy that the performance of the rejection sampling heavily depends on the area ratio between the surface inside the depth contours and the total surface within the bounding box. As the area ratio decreases, the probability of successfully computing a valid sample also decreases, leading to slower sampling performance. This is clearly shown in Fig.~\ref{fig:sampling_performance} where an area ratio difference of about 14\% between the Great Belt and Little Belt study-cases produces a change of about 40\% in computational time. This issue is mitigated by utilizing the computed triangulation, since it directly represents the valid sampling space, guaranteeing that any computed sample is valid, resulting in no wasted sampling effort.

\section{Conclusion}
The paper presented a grounding-aware RRT* path planning algorithm for collision avoidance in confined waters. The proposed algorithm combines the navigation experience of seafarers with the water depth information available through the ENC, to achieve optimal path alterations that trade off between comfortable navigation (not too close to shallow waters) and feasible navigation (crossing navigable areas for the current draught). The paper proposed the use of multivariate kernel density estimation on historical AIS data to generate a probabilistic model that describes the best sailing practices of seafarers in a given restricted sea region. This model is then used as cost factor in the performance index of a COLREGs compliant  RRT* collision avoidance scheme. This enables the planner to compute safe path deviations, which are aware of both collision and grounding risks. Furthermore, compared to existing marine-oriented RRT* implementations, the presented scheme directly utilizes the complex geometries present within the chart data. Through triangulation the sampling scheme has increased probability of improving the computed solution. The augmented planning scheme was successfully demonstrated in simulation for two separate confined water collision scenarios, where the results clearly show the value provided by taking advantage of the prior navigational experiences.

\bibliography{ifacconf}
\end{document}